\documentclass[10pt,twocolumn,letterpaper]{article}

\usepackage{cvpr}
\usepackage{times}
\usepackage{epsfig}
\usepackage{graphicx}
\usepackage{amsmath}
\usepackage{amssymb}
\usepackage{array}
\usepackage{url}

 \usepackage{float}

\usepackage[draft,inline,nomargin]{fixme}
\usepackage{soul}

\usepackage[pagebackref=false,breaklinks=true,letterpaper=true,colorlinks,bookmarks=false]{hyperref}

\cvprfinalcopy 

\def\cvprPaperID{3795} 

\ifcvprfinal\fi
\begin{document}

\title{Learning Latent Super-Events to Detect Multiple Activities in Videos}

\author{AJ Piergiovanni and Michael S. Ryoo\\
Department of Computer Science, Indiana University, Bloomington, IN 47408 \\
\texttt{\{ajpiergi,mryoo\}@indiana.edu}\\
}

\maketitle
\thispagestyle{empty}

\begin{abstract}
In this paper, we introduce the concept of learning latent \emph{super-events} from activity videos, and present how it benefits activity detection in continuous videos. We define a super-event as a set of multiple events occurring together in videos with a particular temporal organization; it is the opposite concept of sub-events. Real-world videos contain multiple activities and are rarely segmented (e.g., surveillance videos), and learning latent super-events allows the model to capture how the events are temporally related in videos. We design \emph{temporal structure filters} that enable the model to focus on particular sub-intervals of the videos, and use them together with a soft attention mechanism to learn representations of latent super-events. Super-event representations are combined with per-frame or per-segment CNNs to provide frame-level annotations. Our approach is designed to be fully differentiable, enabling end-to-end learning of latent super-event representations jointly with the activity detector using them. Our experiments with multiple public video datasets confirm that the proposed concept of latent super-event learning significantly benefits activity detection, advancing the state-of-the-arts.
\end{abstract}

\section{Introduction}
Activity detection is an important computer vision problem with many societal applications, including smart surveillance, monitoring of patients or elderly (e.g., for quality-of-life systems), online video retrieval, and robot perception. Given a continuous video, the task is to find the frames corresponding to every event occurring in the video. This is more challenging compared to the activity classification problem of categorizing a pre-segmented video or localizing a single activity in a trimmed video. Although activity detection is an important area to study as almost all real-world videos contain multiple activities and are rarely segmented (e.g., surveillance systems), it has been investigated much less, particularly for multi-event videos.

In the past years, end-to-end learning methods using convolutional neural networks (CNNs) obtained a great amount of success in video analysis. These approaches successfully modeled per-frame (or per-local-segment) information in activity videos, such as a single RGB frame or optical flows over a small number of frames \cite{feichtenhofer2016convolutional}. Recently, models such as I3D~\cite{carreira2017quo} have been developed to capture longer-term dynamics (e.g., 64 frames). However, because such end-to-end models are optimized for capturing per-segment information, the primary focus of existing works has been mainly on activity classification and not detection. There are recent works on activity detection using end-to-end models (e.g., the detection task in \cite{sigurdsson2016hollywood}), but many of these works also focus on making better per-segment decisions (and their post-processing) rather than learning details of temporal structure/context over the entire (variable length) video.

\begin{figure}
  \centering
    \includegraphics[width=1.0\linewidth]{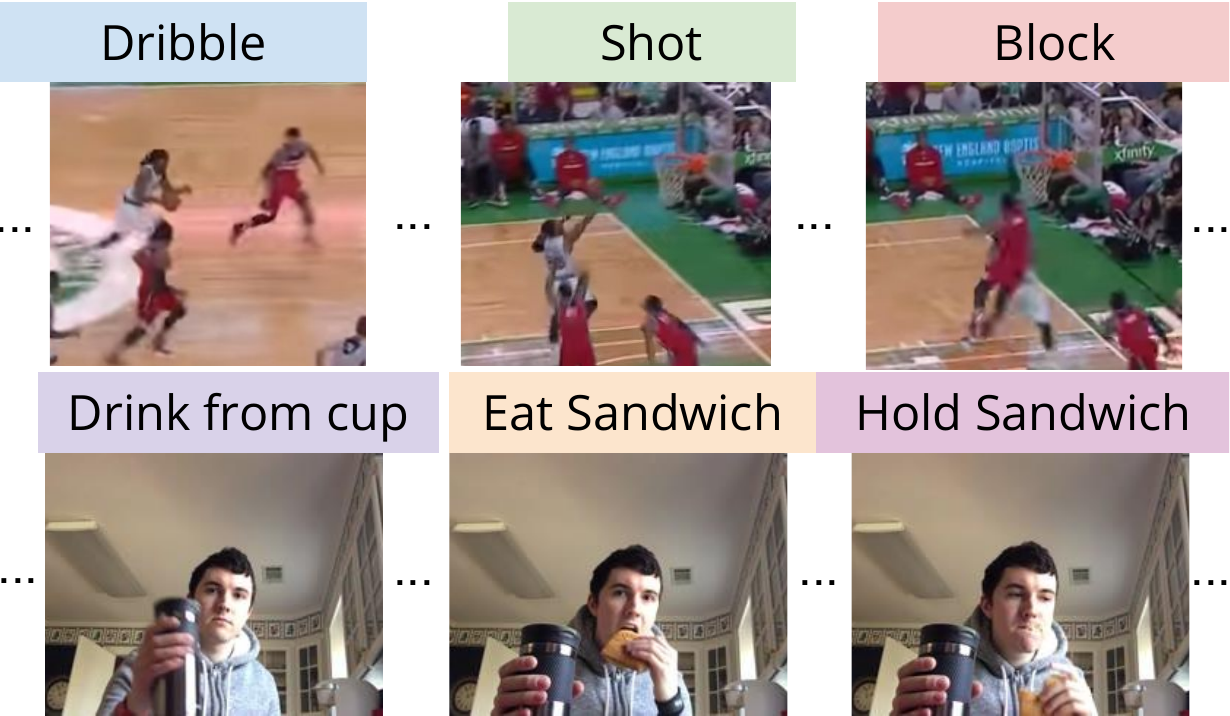}
      \caption{{\bf Top:} In a basketball video, the block event cannot occur without a shot event. Similarly, the dribble event provides temporal context as to where a shot event can occur. {\bf Bottom:} In a video of a person eating, `eating a sandwich' must occur near `holding sandwich'. The drinking action is also related to eating.}
      \label{fig:temporal-example}
\end{figure}

\begin{figure*}
  \centering
    \includegraphics[width=\textwidth]{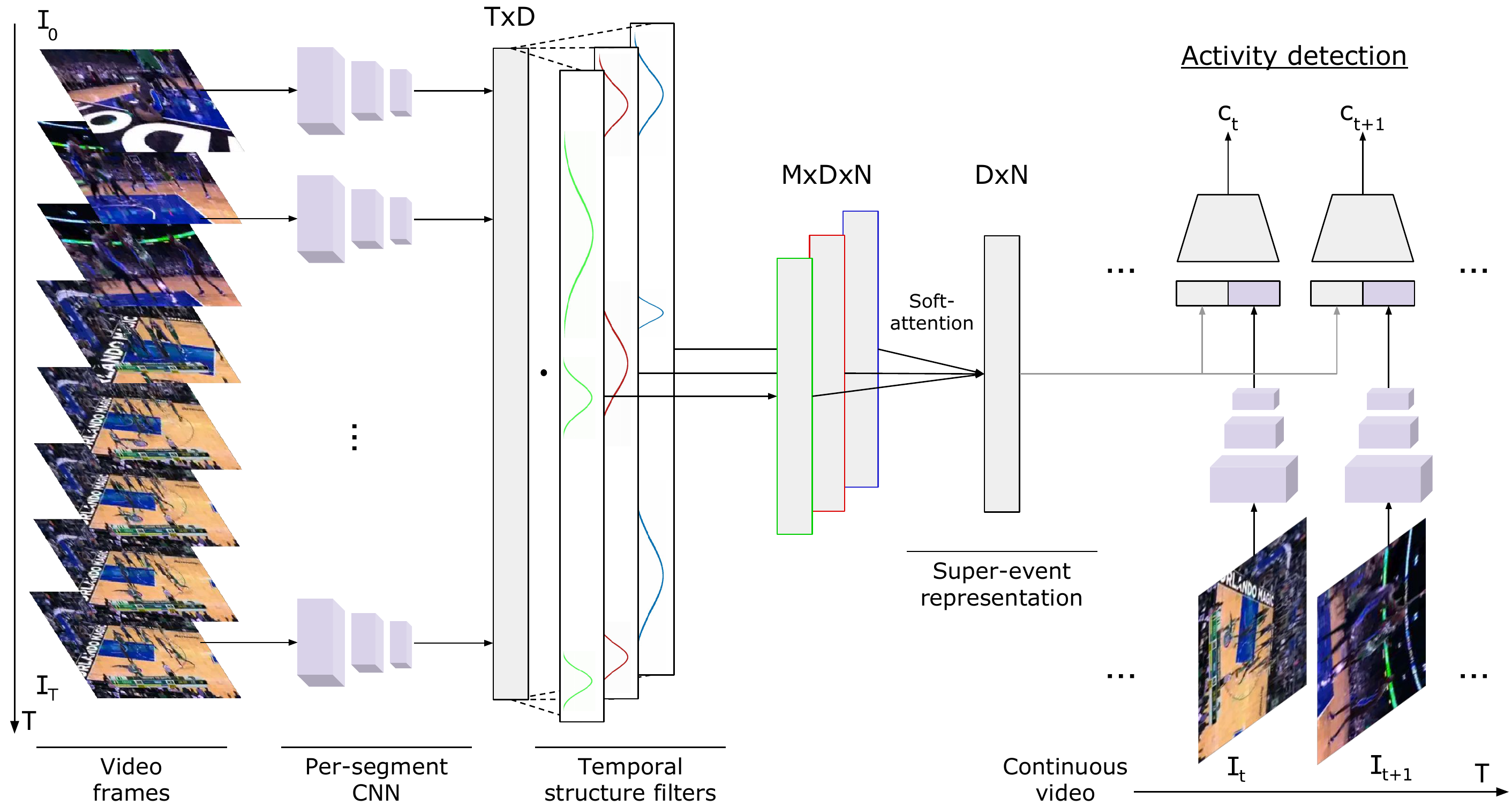}
      \caption{Overview of our approach. The temporal structure filters are applied to the entire video and soft-attention is applied to form a \emph{super-event} representation for each video. We concatenate per-frame (or per-segment) video features with the super-event representation to make per-frame classifications, annotating each frame with its activity class (or no-activity).}
      \label{fig:model-overview}
\end{figure*}

A video contains multiple activities (in a sequence or in parallel) and they are correlated. This means that detecting the frames of one activity in the video should benefit from information in the frames corresponding to another activity, which are often temporally very separated. Existing approaches of representing/classifying video segments without regard to contextual information is thus limited; continuous videos contain rich temporal structure which can be exploited to improve activity detection. For example, in a video of a basketball game, shooting and blocking events must occur near-by, as shown in Figure~\ref{fig:temporal-example}. A block or rebound event cannot occur without a shot event.


In this paper, we introduce the concept of \emph{super-events} and present how its representation learning can benefit activity detection. We define a super-event as a set of multiple events occurring together in videos with a particular temporal pattern (i.e., structure). More specifically, if the event we are interested in is a part of a longer-term event, we call the longer-term event its super-event. This is the opposite concept of `sub-events'. For example, the events of shooting and blocking mentioned above forms a super-event, which may be named as a blocked-shoot. Learning such latent super-events allows the model to capture how the events are temporally related in their videos. Once learned, when making a prediction (i.e., testing), the super-events can serve as temporal context to better detect the events. This enables the detection decision at each frame to be made while considering longer-term temporal structure. Note that such super-events are `latent', meaning that no super-event annotations are provided.

We newly design \emph{temporal structure filters}, and convolve it with the video representation to obtain a super-event representation. Temporal structure filters allow the model to focus on particular sub-intervals. Such temporal structure filters are learned for each event, optimized based on the training data for the best super-event representation construction. For each frame, we combine the super-event representation with the per-frame or per-segment CNN representation for its binary classification per event.  Our method is fully-differentiable and can be learned end-to-end using back propagation, making it suitable for any length video. Our experimental results with three different datasets confirm the benefits of our latent super-event learning, obtaining the state-of-the-art results on MultiTHUMOS and Charades.




\section{Related works}

Activity recognition has been a popular research topic in computer vision~\cite{aggarwal11}. Hand-crafted features, such as improved dense trajectories~\cite{wang2011action} gave excellent results on many benchmark datasets in the past. Recently, there have been more works on learning features for action recognition using convolutional models. Two-stream CNN approaches take both RGB frames and optical flow as input~\cite{simonyan2014two} to capture motion and image features. 3D spatio-temporal (XYT) convolutional filters also have been learned and applied to many activity recognition tasks~\cite{tran2014c3d, carreira2017quo}. Large scale datasets and challenges such as THUMOS~\cite{THUMOS14}, ActivityNet~\cite{caba2015activitynet} and Charades~\cite{sigurdsson2016hollywood} provided these approaches training videos to learn the model.


Improving activity recognition by temporally aggregating such per-frame CNN features has also been studied. Ng et al. \cite{ng2015beyond} compared several different forms of temporal pooling over per-frame CNN features. Karpathy et al. \cite{karpathy2014large} compared various methods to combine temporal information within a CNN. Piergiovanni et al. \cite{piergiovanni2017learning} studied learning multiple sub-intervals to improve activity recognition. These works focused mostly on pooling short-term temporal information to classify a single interval as an activity. They did not detect multiple activity instances or explore long-term structure between multiple activities. 

Recurrent neural networks (RNNs) such as long short-term memory (LSTM) have been popularly used to model temporal event transitions between frames~\cite{yeung2016end,escorcia2016daps,yeung2015every}. These approaches all relied on a single RNN to capture temporal information in different frames, only implicitly capturing relationships between multiple activities. Our approach allows for more explicit modeling of activity relationships as their super-events, which leads to better performance and more insights into what the network learns.

Recently, segment-based 3D CNNs have been used to capture spatio-temporal information simultaneously for the activity detection~\cite{shou2016temporal,xu2017r}. Shou et al. \cite{shou2017cdc} uses convolutional upsampling to make dense predictions to better localize start and end times. Zhao et al. \cite{zhao2017temporal} use a binary actionness classifier to propose many action segments per video then classify each segment individually. However, these approaches all treated the segments as individual instances and do not exploit the longer-term relationships between actions.

Sigurdsson et al. \cite{sigurdsson2017actions} found that using intent, which they defined as the clustering of similar activities, is helpful to activity detection. Fully-connected CRFs were applied as a post-processing of per-frame CNN features as well as object features. 
While such approach learns some global context, they do not learn the explicit temporal structure, nor are they learned in an end-to-end fashion.

To our knowledge, this is the first work exploring an end-to-end model for super-event representation learning, capturing temporal structure and relationships between activities. Learning hierarchical structures of activities has been studied in many traditional works~\cite{niebles2010modeling,gaidon2011actom,laxton2007leveraging,gupta2009understanding,brendel2011learning,ryoo13}, but they were not learned end-to-end or required additional labels for intervals. Our model is fully differentiable, enabling joint end-to-end learning of latent super-events and the activity detector.


\section{Activity detection with latent super-events}

The objective of our model is to annotate each frame as its corresponding activity class (including no-activity) given a continuous video. We present an end-to-end learning model that does such labeling not only by looking at each frame or each local segment but also by considering the overall temporal structure. Our idea is to allow the model to learn the representations of \emph{super-events} summarizing much longer-term temporal intervals, and take advantage of them for the activity detection. A super-event is defined as a longer-term event containing the event of interest, which is the opposite concept of `sub-events'. Our approach, shown in Fig.~\ref{fig:model-overview}, effectively represents super-events using temporal structure filters. We learn per-class soft-attention weights over the filters to create a super-event representation, and take advantage of it for the frame classification.



\subsection{Per-frame representation}
\label{subsec:per-frame}

The base component of our detection is a CNN providing per-frame (or per-local-segment) representation. This is obtained by learning standard video CNN models (e.g., \cite{simonyan2014two,carreira2017quo}). We train the model to learn binary per-frame classifiers by optimizing:
\begin{equation}
    L(v) = \sum_{t,c} z_{t,c}\log(p(c|v_t)) + (1-z_{t,c})\log(1-p(c|v_t))
\end{equation}
where $v_t$ is the per-frame or per-segment CNN feature at frame $t$ and $z_{t,c}$ is the ground truth label for class $c$ and time $t$. This gives a sequence of probabilities for each class which can be used to find activity intervals. Using a fully-connected network to model $p(c|v_t)$ captures minimal temporal information, using just a single frame or segment. RNN models have been used to compute $p(c|v_t)$ which captures some implicit temporal information. 
The learned CNN producing $v_t$ serves as our base component.


\subsection{Temporal structure filter}
\begin{figure}
  \centering
    \includegraphics[width=\linewidth]{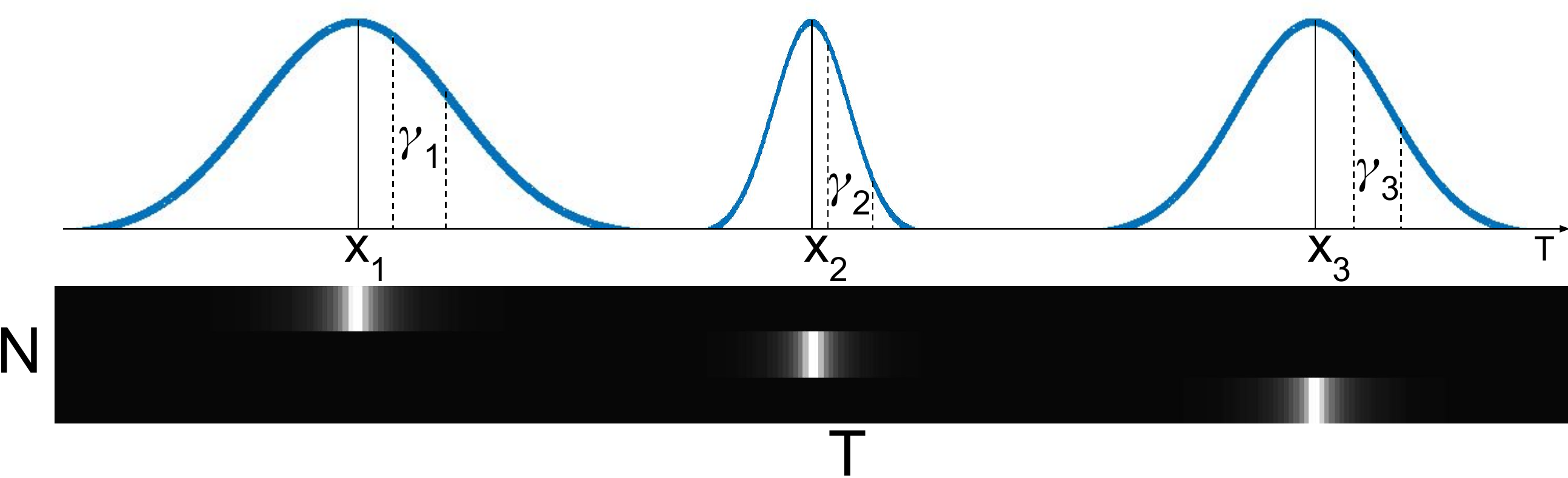}
      \caption{An illustration of a temporal structure filter. Each filter learns independent centers and widths.}
      \label{fig:temporal-filter}
\end{figure}

The \emph{temporal structure filter} we introduce in this paper is a filter designed to capture temporal context formed by multiple activities. It is an extension/generalization of the spatial attention model proposed in~\cite{gregor2015draw}. The idea is to represent a variable length video with a fixed dimensional vector, by only focusing on the `learned' frame locations. 
The previous attention model repeats a single Gaussian distribution several times with a fixed stride, while a temporal structure filter learns several independent distributions.

In particular, each temporal structure filter is modeled as a set of $N$ Cauchy distributions: we found that the Cauchy distribution was easier to train than the Gaussian distribution commonly used (i.e., it converges faster). Each distribution learns a center, $x_n$ and $\gamma_n$ which controls the width. Given $T$, the length of the video, each filter is constructed by:
\begin{equation}
\begin{split}
  \hat{x_n} &= \frac{(T - 1) \cdot (\tanh\left(x_n\right)+1)}{2}\\
  \hat{\gamma_n} &= \exp(1 - 2\cdot|\tanh\left(\gamma_n\right)|)\\
  F[t,n] &= \cfrac{1}{Z_n\pi\hat{\gamma_n}\left(\cfrac{(t-\hat{x_n})}{\hat{\gamma_n}}\right)^2}
\end{split}
\end{equation}
where $Z_n$ is a normalization constant, $t\in\{1,2,\ldots, T\}$ and $n\in\{1,2,\ldots,N\}$. Figure~\ref{fig:temporal-filter} shows an example.

When trained for the super-event representation learning, our temporal structure filter allows the model to explicitly learn which temporal intervals in the entire video are relevant for the frame-level event detection.

\subsection{Super-event representation learning}
\label{subsec:super}

In its elementary form, we represent the super-event of each event $c$ by applying one temporal structure filter $F_c$ over the entire video representation $v$. This essentially is a matrix product between each of the $N$ distributions in $F_c$ and $v$. The super-event representation $S_c$ is obtained as:
\begin{equation}
    S_c[n] = \sum_t^T F_c[t,n] \cdot v_t.
\end{equation}
This operation applies $F_c$ (an $T\times N$ filter) to $v$ (the $T\times D$ video features) and returns an $N\cdot D$-dimension vector, which we call the super-event representation. This has an effect of summarizing the entire video representation $v$ into a much smaller representation $S_c$ by focusing on the frames specified with the learned parameters of $F_c$.


\vspace{-3pt}
\paragraph{Super-event representation with soft attention:}

As several activities can share the same super-event, it makes sense to learn a set of $M$ different temporal structure filters and share these filters across the classes. Here, $M$ is less than the number of classes $C$. In order to represent the super-event of each activity class $c$ using such $M$ filters, we learn a set of per-class soft-attention weights allowing each activity class to select some of the $M$ structure filters to use. For a set of $C$ classes, we learn weights $W_{c,m}$ and compute the soft-attention as:
\begin{equation}
    A_{c,m}=\cfrac{\exp(W_{c,m})}{\sum_k^{M}\exp(W_{c,k})}.
\end{equation}
We then create a super-event representation by applying these weights to the $M$ temporal structure filters:
\begin{equation}
    S_c = \sum_m^{M} A_{c,m}\cdot \sum_t^T F_{m}[t] \cdot v_t
\end{equation}


\subsection{Detection with super-events}
We perform per-frame binary classification for each class by concatenating the super-event representation with the CNN frame representation:
\begin{equation}
    p(c|[v_t, S_c]) = \sigma(W [v_t, S_c])
\end{equation}
where $W$ is a learnable parameter and $\sigma$ is the sigmoid function. Unlike standard per-frame classification approaches, each frame now depends on the abstracts from the entire video features (i.e., our super-event representation) instead of a single frame.

To learn parameters, we optimize the multi-label binary classification loss:
\begin{equation}
\begin{split}
    L(v) = &\sum_{t,c} z_{t,c}\log(\sigma(W [v_t, S_c])) \\
        &+ (1-z_{t,c})\log(1-\sigma(W [v_t, S_c])
\end{split}
\end{equation}
where $z_{t,c}$ is the ground truth label for class $c$ at time $t$. We minimize this loss using stochastic gradient descent.

\subsection{Relative super-events}
We also propose a relative super-event model, where instead of computing the super-event representation for the entire video, we compute a super-event representation relative to the current frame location. This allows us to capture relative relationships like in basketball, shooting and blocking must occur together, regardless of the other content in the video. This works well even when a video contains multiple, unrelated activity classes, for example a highlight video of various sports.
This approach is identical to the one above, except the length $L$ of the temporal structure filter is fixed. We use the temporal structure filters $F$ as a convolutional kernel and convovle with the video features, applying $F$ centered at each feature $t$ with a length of $L$ frames. The per-class attention weighting is applied to the structure filters to form a per-frame super-event representation used for the classification of the current frame.

\section{Experiments}
In order to evaluate the effectiveness of our proposed super-event representation learning, we conducted a set of experiments comparing our method to conventional approaches across various datasets and feature types. We particularly focused on activity detection datasets with videos containing multiple actions/activities, including MultiTHUMOS~\cite{yeung2015every}, Charades~\cite{sigurdsson2016hollywood}, and AVA~\cite{ava2017}.

\subsection{Settings}

\paragraph{Features:}
We extracted I3D~\cite{carreira2017quo} features from the videos at 25fps with a stride of 8 frames. We are using I3D as the base per-segment CNN mentioned in Section \ref{subsec:per-frame}. I3D is a two-stream 3D CNN that achieved state-of-the-art performance on several action recognition tasks. Using both RGB and optical flow as input, it is able to capture relatively longer-term temporal relationships using a temporal resolution of up to 99 frames. We obtained pre-trained weights for I3D as well as code to fine-tune I3D from the authors, which is identical to what they used for the Charades Challenge 2017. We also tested VGG RGB~\cite{simonyan2014very}, VGG optical flow, and standard two-stream CNNs~\cite{feichtenhofer2016convolutional} at 8fps as our base per-frame CNNs. The optical flow and two-stream features have a temporal resolution of 10 frames and RGB features have a temporal resolution of 1 frame.

\vspace{-3pt}
\paragraph{Implementation Details:} We implemented the temporal structure filters in PyTorch. The learning rate was set to 0.1 and reduced by a factor of 10 every 1000 iterations. We trained the network using a batch size of 32 videos for 5000 iterations using the Adam~\cite{kingma2014adam} optimizer with default parameters. We applied dropout with a probability of 0.5 to the input features. We set $N=3$ (i.e., 3 Cauchy distributions per temporal structure filter), and $M=5$ (i.e., 5 temporal structure filters). Per-frame CNN representation dimensionality is $D=1024$ for I3D features and $D=4096$ for VGG features. In both cases, we used the output of the layer before the final fully-connected layer as the features. In order to make the final per-frame decisions based on per-segment CNN features and our super-event representations, we trained a single fully-connected layer with input size $D + D\cdot N$ and output size of the number of classes. Our code and pretrained models are available at \href{https://github.com/piergiaj/super-events-cvpr18}{https://github.com/piergiaj/super-events-cvpr18}.

\subsection{MultiTHUMOS}

\paragraph{Dataset:}
MultiTHUMOS~\cite{yeung2015every} is an extension of the THUMOS~\cite{THUMOS14} dataset with the untrimmed videos densely annotated for 65 different action classes. Unlike ActivityNet and THUMOS, MultiTHUMOS has on average 10.5 activity classes per video, 1.5 labels per frame, and up to 25 different activity instances in each video. This allows us to confirm the value of learning super-event representations in complex videos. MultiTHUMOS contains YouTube videos of various sport activities, such as basketball games (shoot, guard, dribble, and block), volleyball games (serve, set, block, and spike), types of weightlifting, throwing, etc.

We followed the standard activity detection evaluation setting of MultiTHUMOS, which is measuring the mean average precision (mAP) by annotating each frame in test videos. We actually used fewer training samples than provided: We only used the 1010 continuous videos for our training, and test on the full 1574 test videos. We did not use the segmented training videos from THUMOS, since super-event learning is not meaningful with trimmed videos. Even without using the full dataset, we were able to outperform the previous approaches. 

\vspace{-3pt}
\paragraph{Results:}

\begin{table}
\caption{Performances of the state-of-the-art methods and our approach on MultiTHUMOS. Our approach meaningfully outperforms all previous results.}
\label{res:multithumos}
\centering
\setlength\extrarowheight{0pt}
\begin{tabular}{c|c}
\hline
 & mAP \\
\hline
Two-stream~\cite{yeung2015every}    & 27.6\\
Two-stream + LSTM~\cite{yeung2015every}          & 28.1\\
Multi-LSTM~\cite{yeung2015every}    & 29.6\\
Predictive-corrective~\cite{dave2017predictive} & 29.7\\
I3D baseline                        & 29.7 \\
I3D + LSTM                          & 29.9 \\
I3D + Temporal Pyramid              & 31.2 \\
I3D + our super-event                    & \bf{36.4} \\
\hline
\end{tabular}
\end{table}


Table~\ref{res:multithumos} shows the performance (mAP) of our approach compared against previously reported results of the state-of-the-art methods. We are able to observe that our method outperforms previous approaches by a significant margin, achieving a mAP of 36.4\%. This is meaningfully higher than our baseline (i.e., I3D) and the previous best approach \cite{dave2017predictive}: 29.7\% vs. 36.4\%. Using the pre-trained I3D \cite{carreira2017quo} model provided the accuracy identical to the previous best reported performance, and our approach of learning and using latent super-events on top of I3D outperformed it by the margin of 6.7\%. Note that this is also higher than using a sequential recurrent model like LSTM on top of the same feature, also by a margin of more than 6\%. In addition, we also tested the method of using a temporal pyramid commonly used in the previous works (e.g., \cite{ryoo2015pooled}) as a super-event representation. Using the temporal pyramid (of level 3) gave us the performance of 31.2 on MultiTHUMOS (while ours was 36.4).

\begin{table}
\caption{Performances of different super-event representations, evaluated using MultiTHUMOS. I3D was used as the base CNN.}
\label{res:multithumos2}
\centering
\setlength\extrarowheight{0pt}
\begin{tabular}{c|c}
\hline
 & mAP \\
\hline
Baseline                        & 29.7 \\
Global max pooling              & 30.0 \\
Global mean pooling             & 30.8 \\
Temporal pyramid pooling        & 31.2 \\
Single super-event (per-class)  & 31.2 \\
With soft attention             & 36.4 \\
With soft attention + relative  & 36.2 \\
\hline
\end{tabular}
\end{table}

We conducted more experiments to compare different super-event representations in our approach. Table~\ref{res:multithumos2} compares (1) the baseline per-frame classifier, (2) the method of using global mean/max pooling as a super-event representation, (3) temporal pyramid pooling, (4) the elementary single super-event representation mentioned in Section \ref{subsec:super}, (5) our soft-attention-based super-event representation, and (6) the relative super-event approach. We found that while the global max/mean pooled representation and the single super-event improves performance (compared to the baseline), using the soft-attention and shared temporal structure filters gave the best performance. Relative super-events performed similarly to non-relative super-events.

\begin{figure*}
  \centering
    \includegraphics[width=\linewidth]{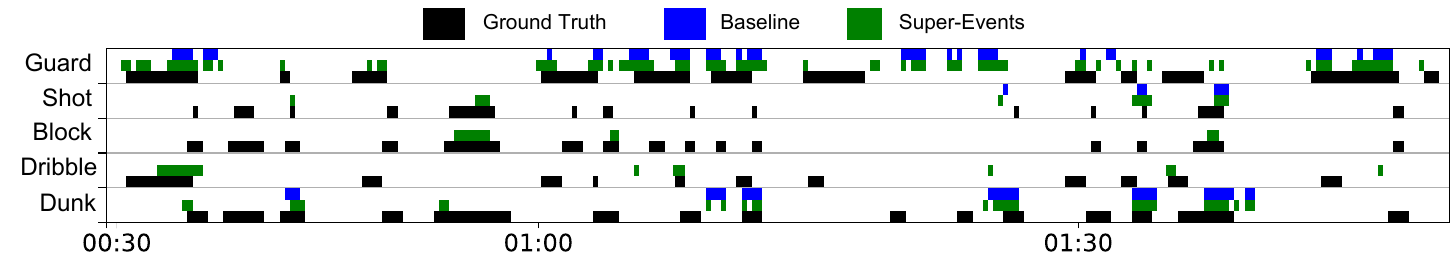}
      \caption{Results from a video in MultiTHUMOS. Super-events especially help the detection of the shooting and blocking events.}
      \label{fig:results-multithumos}
\end{figure*}

\begin{figure*}
  \centering
    \includegraphics[width=\linewidth]{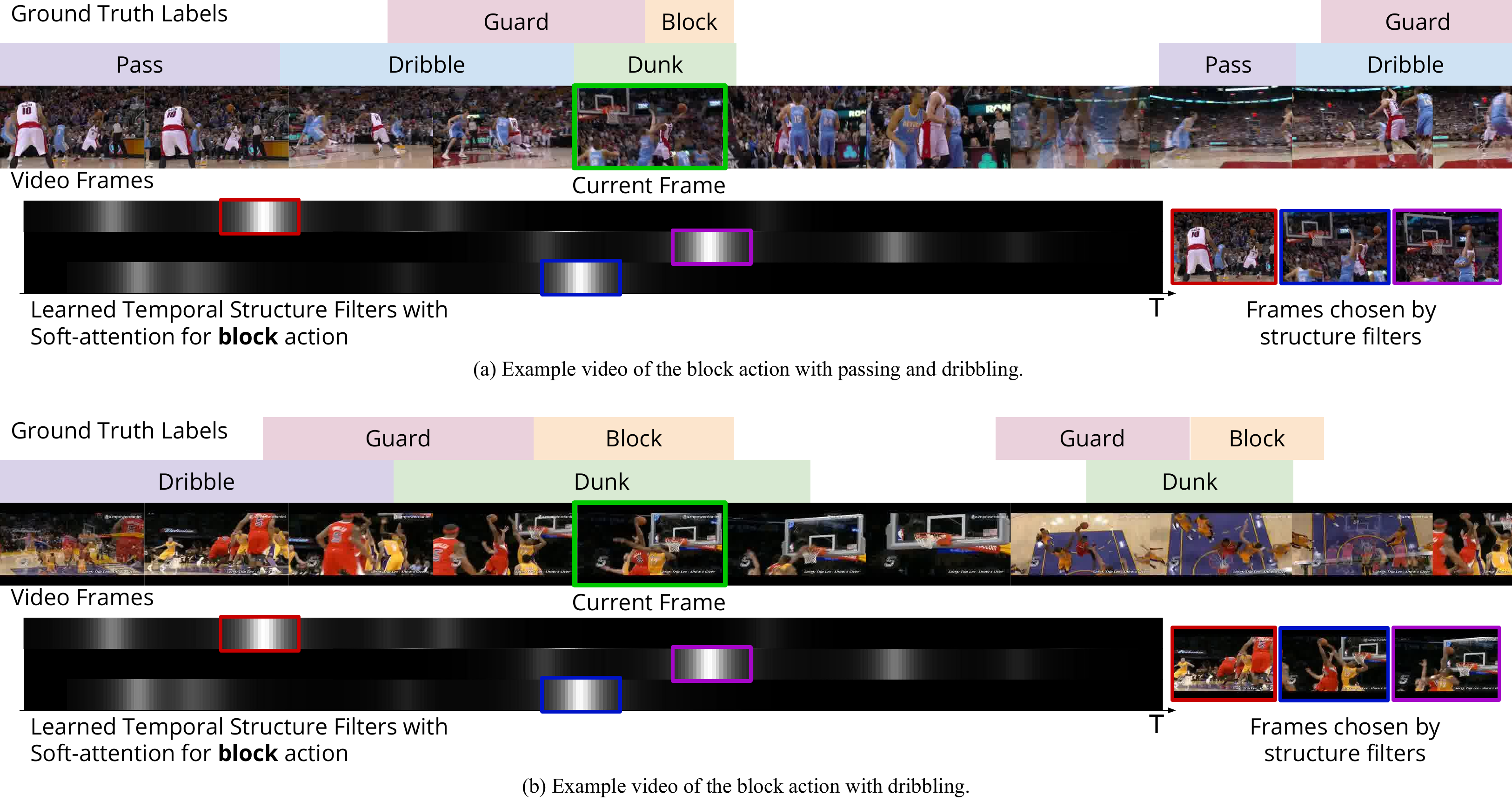}
      \caption{Illustration of the learned temporal structure filters with soft-attention for the block action in MultiTHUMOS. When applied to two different videos, we can observe that the temporal structure filters capture the temporal relationships between the frames corresponding to shooting and blocking action, as well as the relationship between those corresponding to dribble/pass and blocking action.}
      \label{fig:learned-multithumos}
\end{figure*}

Figure~\ref{fig:results-multithumos} illustrates a qualitative analysis of our results. We can observe that super-events improve the detection of related basketball events, especially the co-occurrence of the shooting and blocking actions. The learned temporal structure filters and super-event representations are shown in Figure~\ref{fig:learned-multithumos}. The filters are visualized by combining our learned temporal structure filters using their soft-attention weights. We obtain 5 TxN filters, and added them based on the learned attention weights.

\begin{figure*}
  \centering
    \includegraphics[width=\linewidth]{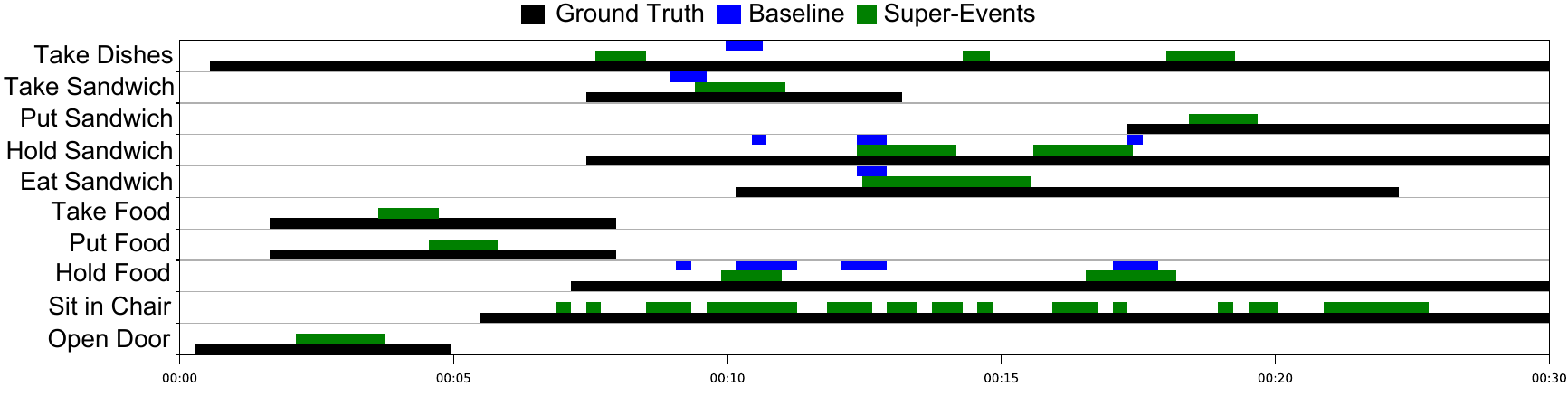}
      \caption{Results from a video in Charades. Super-events improve the detection of related events such as taking, holding and eating a sandwich.}
      \label{fig:results-charades}
\end{figure*}

\subsection{Charades dataset}

\paragraph{Dataset:}
Charades~\cite{sigurdsson2016hollywood} is a large scale dataset consisting of 9848 videos across 157 activities. The videos were recorded by people in their own homes based on a provided script. Each video contains on an average of 6.8 activity instances, often with complex co-occurring activities, making it a suitable dataset to test our super-event learning. The activities were mainly performed at home, and their classes include `preparing a meal', `eating', `sitting', `cleaning', etc. 

In our experiments, we follow the original Charades test setting (i.e., Charades\_v1\_localize evaluation). This is a bit different from the newer test set released for the Charades Challenge 2017. The ground truth labels of the challenge test videos are not publicly available and its evaluation server is not approving any new account access. Instead, we followed the original Charades localization test setting (v1) from the dataset website. The original setting is more challenging than the competition setting in the aspect that the competition allowed more training data to be used than the original setting: the participants were able to use the videos of the original test set as a part of the training in the competition. Note that the exact same I3D code (with fine-tuning) that provided 20.72 mAP in the competition setting is only providing 17.22 mAP in the original test setting.


Similar to MultiTHUMOS, the performances are measured in terms of mAP by evaluating per-frame annotations.


\begin{table}
\caption{Results comparing our approach using super-events with the baselines and LSTMs on the Charades dataset v1. These numbers are raw results without the post-processing method of \cite{sigurdsson2016asynchronous}.}
\label{res:charades-exps}
\centering
\setlength\extrarowheight{0pt}
\begin{tabular}{c|c|c|c}
\hline
 & Baseline & ~LSTM~ & ~~Ours~~ \\
\hline 
\multicolumn{1}{c}{Two-stream CNN}\\
\hline
VGG RGB                 & 6.11 & 6.23 & 7.64 \\
VGG Flow                & 5.13 & 6.08 & 6.83 \\
VGG Two-Stream          & 6.56 & 7.85 & 8.53 \\
\hline 
\multicolumn{1}{c}{Pre-trained I3D}\\
\hline
I3D RGB   & 9.78 & 9.91 & 11.0 \\
I3D Flow  & 9.58  & 9.81 & 11.0 \\
I3D Two-Stream          & 10.32 & 11.50 & 12.8\\
\hline
\multicolumn{1}{c}{Fine-tuned I3D}\\
\hline
I3D RGB      & 15.63 & 17.03 & 18.64 \\
I3D Flow     & 16.21 & 17.54 & 18.52 \\
I3D Two-Stream          & 17.22 & 18.12 & 19.41 \\
\hline
\end{tabular}
\end{table}

\vspace{-3pt}
\paragraph{Results:}

We compared the effectiveness of our approach of using super-event representations with multiple different base per-frame/per-segment CNNs. More specifically, we used the I3D features across RGB frames, flow frames, and two-stream with and without fine-tuning on the Charades dataset. We also tested our approach with the standard two-stream CNN using VGG models. These features capture various amounts of temporal information, from a single RGB frames (VGG RGB), to 10 optical flow frames (VGG Flow and VGG two-stream) and up to 99 frames for I3D.

Table~\ref{res:charades-exps} shows the results describing how much our latent super-event learning approach improves the activity detection performance for each per-frame/per-segment baseline. We are reporting the performances of our method using the super-event representations with soft attention. Furthermore, we implemented the standard LSTM method over the same baseline CNNs. The idea was to directly compare the abilities of our approach and the LSTM in capturing long-term temporal dynamics/relations. We found that super-events yield meaningful performance increases regardless of the feature type. We also outperform the LSTM models, confirming that super-events are better able to capture and use temporal structure than LSTMs.

\begin{figure*}
  \centering
    \includegraphics[width=\linewidth]{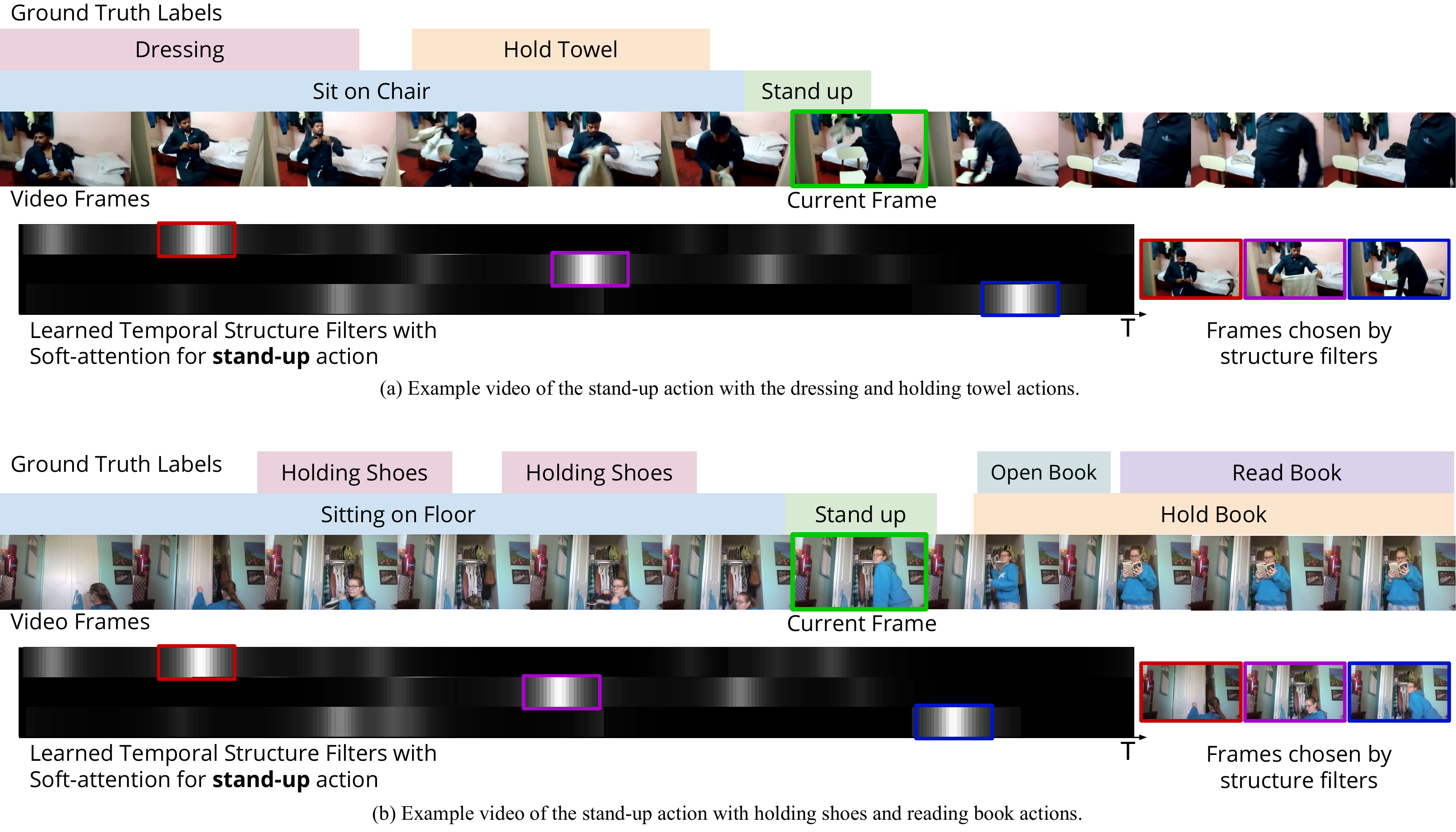}
      \caption{Illustration of the learned temporal structure filters with soft-attention for the stand-up action in Charades. The filters are applied to two different videos. It shows that our temporal structure filters are able to capture the relations that the frames of a sitting action come before the frames of standing up, regardless of the other actions that may occur in the video.}
      \label{fig:learned-charades}
\end{figure*}

We compare our results with the state-of-the-art in Table~\ref{res:charades-comp}. Our method is obtaining the best known performance in the localization setting of the Charades dataset. Notably, it is performing better than I3D which obtained the best competition performance in 2017, while using the same feature. Figures~\ref{fig:results-charades} and \ref{fig:learned-charades} show example detections.


\begin{table}
\caption{Results on Charades original dataset (i.e., Charades\_v1\_localize setting). Note that this setting is a bit different from the Charades Challenge 2017 competition setting, whose evaluation server is not approving any new account access. This setting uses less training data.}
\label{res:charades-comp}
\centering
\setlength\extrarowheight{0pt}
\begin{tabular}{c|c}
\hline
 & mAP \\
\hline
Random~\cite{sigurdsson2016asynchronous} & 2.42 \\
RGB~\cite{sigurdsson2016asynchronous}    & 7.89 \\
Predictive-corrective~\cite{dave2017predictive} & 8.9 \\
Two-Stream~\cite{sigurdsson2016asynchronous}& 8.94 \\
Two-stream+LSTM~\cite{sigurdsson2016asynchronous}& 9.6 \\
R-C3D~\cite{xu2017r}                     &  12.7 \\
Sigurdsson et al.~\cite{sigurdsson2016asynchronous} & 12.8 \\
I3D~\cite{carreira2017quo}                & 17.22 \\
I3D + LSTM                               & 18.1 \\
I3D + Temporal Pyramid                    & 18.2 \\
I3D + super-events                       & \bf{19.41} \\
\hline
\end{tabular}
\end{table}

\subsection{AVA dataset:}

\paragraph{Dataset:}

AVA~\cite{ava2017} is a new large-scale video dataset containing 80 action classes in 57,600 video clips drawn from 192 movies. Unlike Charades, which has individual activity classes per objects, the activities in AVA are very generic, such as sit, stand, walk, carry, etc.  A 15 minute segment is selected from each movie and annotated in 3 second intervals for spatial and temporal activity detection. Since we are interested in temporal activity detection, we designed a new setting using AVA in which we label each frame with the activities that are occurring in it regardless of spatial location. Identical to Charades (and MultiTHUMOS), we evaluate our models using per-frame mean average precision (mAP). Because each 3 second interval contains the same labels, we average our predictions and produce one probability vector per 3 second interval. We only use temporal annotations in this experiment.

\vspace{-3pt}
\paragraph{Results:}

Table~\ref{res:ava} compares our approach with random, I3D baseline, and I3D + LSTM. We used RGB, Flow, and Two-stream versions of I3D.
Again, we are reporting the performances of our method using super-event representations with soft attention. The results are very consistent with the results we obtained from MultiTHUMOS and Charades. Learning (latent) super-event representations with our method and using them for the activity detection always achieved the higher performance. Using LSTM was also able to improve the vanilla baseline, but our approach always outperformed the LSTM by a meaningful margin.



\begin{table}
\caption{Results on Ava Test Set (mAP), with the temporal annotation setting. That is, we evaluated the accuracy solely based on frame-level annotations.}
\label{res:ava}
\centering
\setlength\extrarowheight{0pt}
\begin{tabular}{c|c|c|c}
\hline
 & Baseline & LSTM & Ours\\
\hline
Random     & 2.65 & 2.65 & 2.65\\
I3D RGB    &  6.8 & 7.0  & 8.3 \\
I3D Flow   &  7.1 & 7.2  & 9.1 \\
I3D Two-Stream & 7.5 & 7.8 & \bf{9.8} \\
\hline
\end{tabular}
\end{table}



\section{Conclusion}

We introduced the concept of learning latent super-events in activity videos. We defined a super-event as a set of multiple events occurring together in videos with a particular temporal structure (i.e., the opposite concept of sub-events). We newly designed the temporal structure filters, and presented how they can be used to capture temporal dynamics in multi-activity videos for the super-event representation learning. We provided a fully differentiable end-to-end architecture to jointly learn the latent super-events and the activity detector using them. We were able to confirm that our method performs superior to the state-of-the-art methods in multiple different activity detection datasets, including MultiTHUMOS and Charades.

\vspace{-3pt}
\paragraph{Ack:}

This work was supported by the U.S. Army Research Laboratory under Cooperative Agreement DAAD 19-01-2-0012, and by the ICT R\&D program of MSIP/IITP, Republic of Korea (17ZF1200, Development of XDMedia Solution for Invigoration of Realistic Media Industry).

{\small
\bibliographystyle{ieee}
\bibliography{egbib}
}

\clearpage
\appendix
\section{Appendix}

\begin{minipage}{\textwidth}%

\begin{figure}[H]
\includegraphics[width=0.99\linewidth]{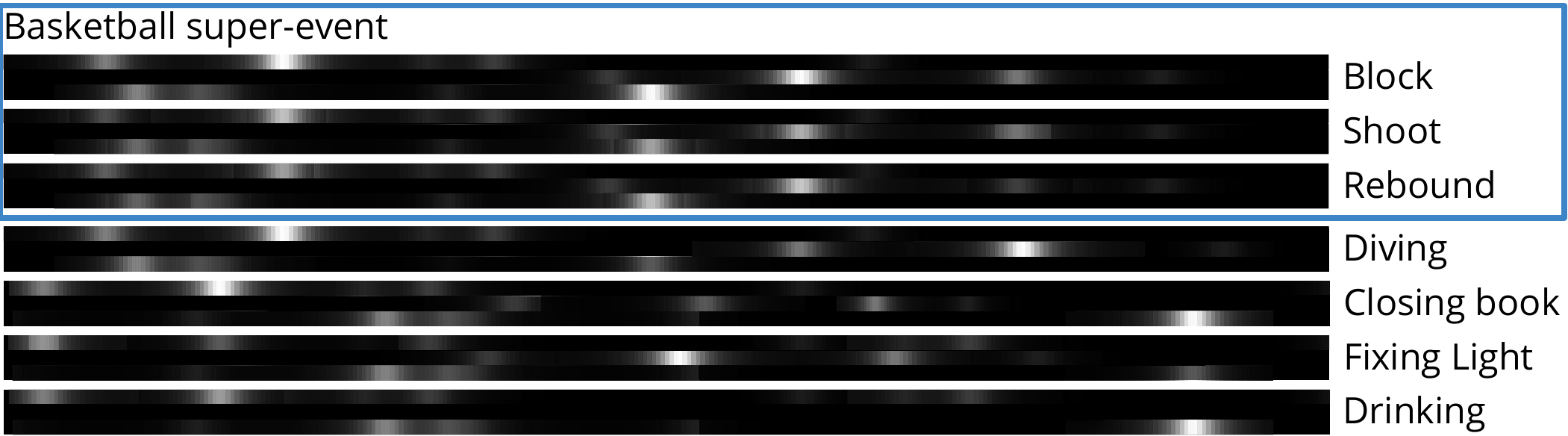}
 \caption{Comparison of the learned temporal structure filters on several different activity classes. The filters are visualized by combining our learned temporal structure filters using their soft-attention weights. We obtain 5 $T\times N$ filters, and sum them based on the learned attention weights. These are global super-event representations which select intervals from the entire video. These filters are scaled to match the length of the video by construction. As a result, even though the videos are continuous and have different lengths, these filters capture the temporal relationships/ordering between the activities assuming their overall relative locations within each video are similar. If such assumption does not hold, we can use the relative version of our super-event representation with more computation. This figure shows that the basketball activities end up learning a very similar global super-event structure (i.e., they share it), while unrelated activities learn different temporal structures.}
\end{figure}

\end{minipage}

\end{document}